\documentclass[10pt,journal,compsoc]{IEEEtran}
\ifCLASSOPTIONcompsoc
  \usepackage[nocompress]{cite}
\else
  \usepackage{cite}
\fi
\ifCLASSINFOpdf
\else
\fi
\usepackage{times}
\usepackage{soul}
\usepackage{url}
\usepackage[hidelinks]{hyperref}
\usepackage[utf8]{inputenc}
\usepackage[small]{caption}
\usepackage{graphicx}
\usepackage{amsmath}
\usepackage{booktabs}
\usepackage{algorithm}
\usepackage{helvet}
\usepackage{courier}
\usepackage{amssymb}
\usepackage{algorithmicx}
\usepackage{algpseudocode}
\usepackage{multirow}
\usepackage{graphicx}
\usepackage{epstopdf}
\usepackage{subfigure}
\usepackage{color}
\usepackage{diagbox}
\usepackage{hyperref}
\usepackage{breakurl}
\usepackage{amsthm}

\newtheorem{theorem}{Theorem}

\newtheorem{definition}{Definition}

\newcommand{\ignore}[1]{}

\usepackage{xcolor}

\begin{document}
%
\title{SecureBoost: A Lossless Federated Learning Framework}
%
%
%
%

\author{Kewei Cheng, Tao Fan, Yilun Jin, Yang Liu,~\IEEEmembership{Member,~IEEE}, Tianjian Chen, Dimitrios Papadopoulos, Qiang Yang,~\IEEEmembership{Fellow,~IEEE}
\thanks{Yang Liu and Qiang Yang are the corresponding authors. Email: yangliu@webank.com, qyang@cse.ust.hk}
\thanks{Kewei Cheng is with the University of California, Los Angeles, Los Angeles, USA. Tao Fan, Yang Liu, Tianjian Chen are with Department of Artificial Intelligence, Webank, Shenzhen, China. Yilun Jin, Dimitrios Papadopoulos are with Hong Kong University of Science and Technology. Qiang Yang is with both Webank and Hong Kong University of Science and Technology.}}

\IEEEtitleabstractindextext{%
\begin{abstract}
The protection of user privacy is an important concern in machine learning, as evidenced by the rolling out of the General Data Protection Regulation (GDPR) in the European Union (EU) in May $2018$. The GDPR is designed to give users more control over their personal data, which motivates us to explore machine learning frameworks for data sharing that do not violate user privacy. To meet this goal, in this paper, we propose a novel lossless privacy-preserving tree-boosting system known as SecureBoost in the setting of federated learning. SecureBoost first conducts entity alignment under a privacy-preserving protocol and then constructs boosting trees across multiple parties with a carefully designed encryption strategy. This federated learning system allows the learning process to be jointly conducted over multiple parties with common user samples but different feature sets, which corresponds to a vertically partitioned data set. An advantage of SecureBoost is that it provides the same level of accuracy as the non-privacy-preserving approach while at the same time, reveals no information of each private data provider. We show that the SecureBoost framework is as accurate as other non-federated gradient tree-boosting algorithms that require centralized data and thus it is highly scalable and practical for industrial applications such as credit risk analysis. To this end, we discuss information leakage during the protocol execution and propose ways to provably reduce it.
\end{abstract}

}

\maketitle

\section{Introduction}
\begin{figure*}[!t]
\centering
\includegraphics[width=1\linewidth]{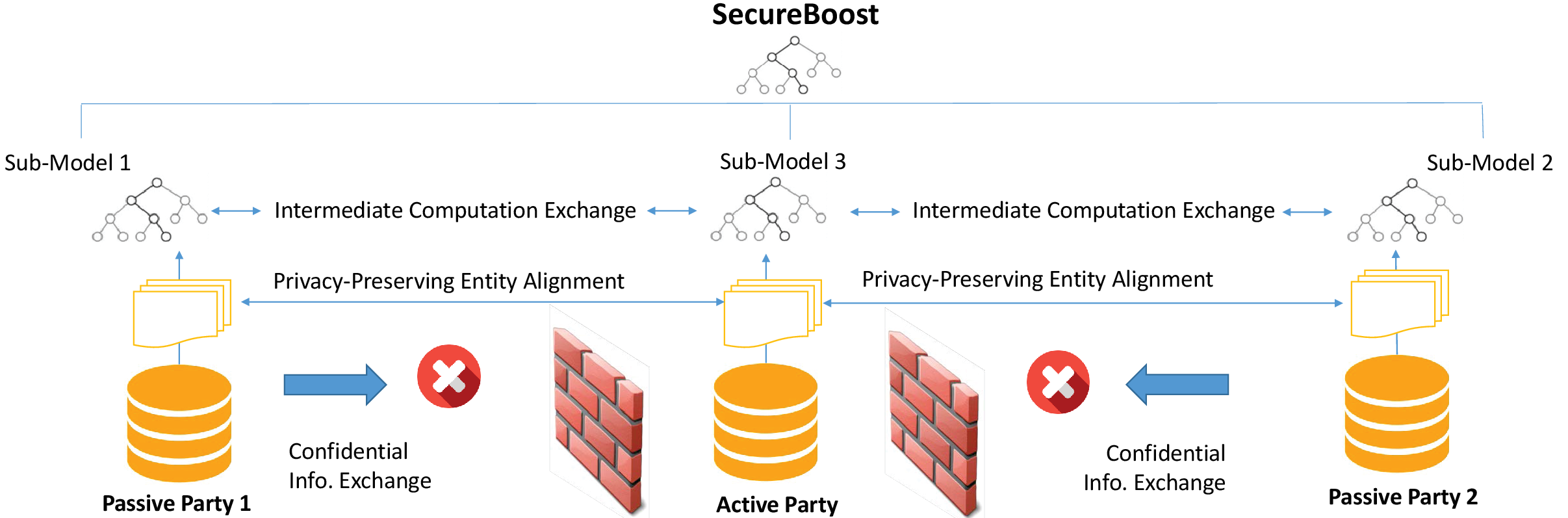}
\caption{Illustration of the proposed SecureBoost framework}\label{Fig.SecureBoost}
\end{figure*}
The modern society is increasingly concerned with the unlawful use and exploitation of personal data.  At the individual level, improper use of personal data may cause potential risk to user privacy. At the enterprise level, data leakage may have grave consequences on commercial interests.  Actions are being taken by different societies.  For example, the European Union has enacted a law known as General Data Protection Regulation (GDPR). GDPR is designed to give users more control over their personal data~\cite{regulation2016general,albrecht2016gdpr,mayer2015regime,goodman2016european}. Many enterprise that rely heavily on machine learning are beginning to make sweeping changes as a consequence.

Despite difficulty in meeting the goal of user privacy protection, the need for different organizations to collaborate while building machine learning models still stays strong. In reality, many data owners do not have sufficient amount of data to build high-quality models. 
For example, retail companies have users' purchases and transaction data, which are highly useful if provided to banks for credit rating applications. Likewise, mobile phone companies have users' usage data, but each company may only have a small amount of users which are not enough to train high-quality user preference models. Such companies have strong motivation to collaboratively exploit the joint data value.

So far, it is still a challenge to allow different data owners to collaboratively build high-quality machine learning models while at the same time protecting user data privacy and confidentiality.  In the past, several attempts have been made to address the user privacy issue in machine learning ~\cite{hardy2017private,mohassel2017secureml}.  For example,  Apple proposed to use {\em differential privacy} (DP)~\cite{dwork2014algorithmic,dwork2008differential} to address the privacy preservation issue. The basic idea of DP is to add properly calibrated noise to data to disambiguate the identity of any individuals when data is being exchanged and analyzed by a third party. However, DP only prevents user-data leakage to a certain degree and cannot completely rule out the identity of an individual. In addition, data exchange under DP still requires that data change hands between organizations, which may not be allowed by strict laws like GDPR.  Furthermore, the DP method is {\em lossy} in machine learning in that models built after noise is injected may perform unsatisfactorily in prediction accuracy.

More recently, Google introduced a {\em federated learning} (FL) framework~\cite{konevcny2016federated} and deployed it on Android cloud.  The basic idea is to allow individual clients to upload only model updates but not raw data to a central server where the models are aggregated.  A secure aggregation protocol was further introduced~\cite{Bonawitz:2017:PSA:3133956.3133982} to ensure the model parameters do not leak user information to the server.  This framework is also referred to as horizontal FL~\cite{Yang-et-al:2019} or data-partition FL where each partition corresponds to a subset of data samples collected from one or multiple users.

\begin{figure}
\centering
\includegraphics[width=1\linewidth]{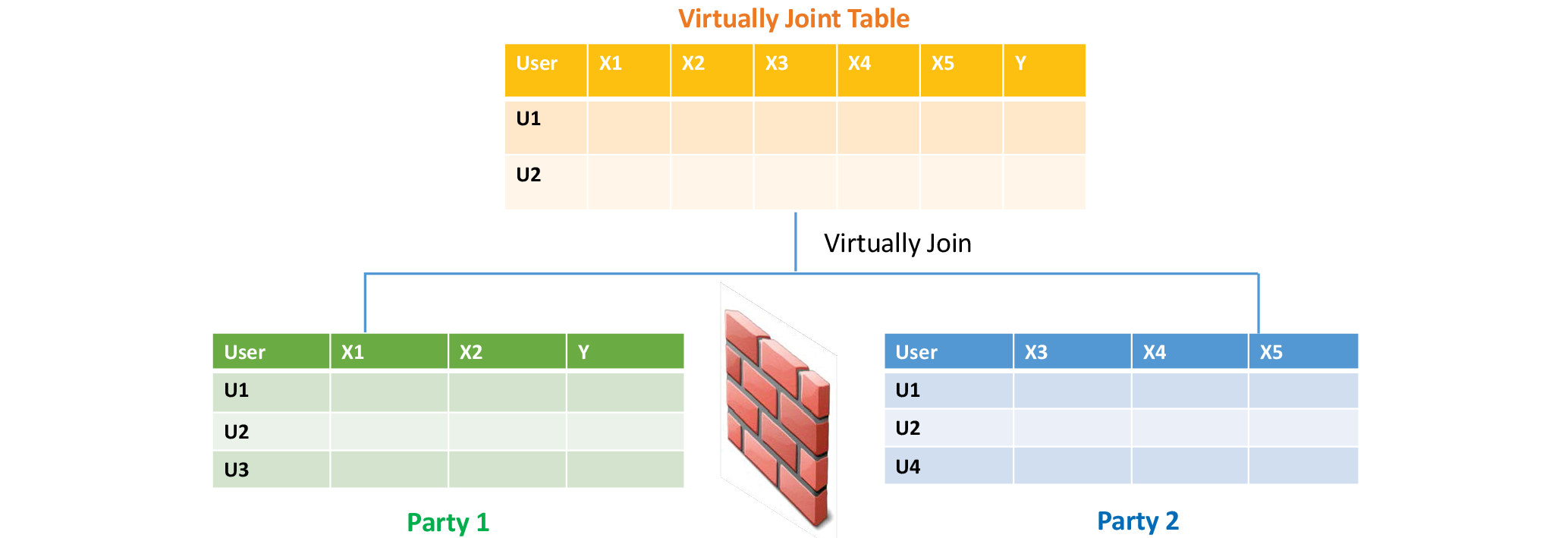}
\caption{Vertically partitioned data set}\label{Fig.Data_Partition}
\end{figure}

In this paper, we consider another setting of multiple parties collaboratively build their machine learning models while protecting user privacy and data confidentiality. Our setting is shown in Figure~{\ref{Fig.Data_Partition}} and is typically referred as vertical FL~\mbox{\cite{Yang-et-al:2019}} because data are partitioned by features among different parties. This setting has a wide range of real-world applications. For example, financial institutes can leverage alternative data from a third party to enhance users' and small and medium enterprises' credit ratings~\mbox{\cite{yang2019federated}}. Patents' record from multiple hospitals can be used together for diagnoses~\mbox{\cite{communication-YangLiu2019,liang2021selfsupervised}}.   
We can regard the data located at different parties as a subsection of a virtual big data table obtained by taking the union of all data at different parties. Then the data at each party has the following property:
\begin{enumerate}
\item The big data table is vertically split, such that the data are split in the feature dimension among parties;
\item Only one data provider has the label information; 
\item Parties share a common set of users.
\end{enumerate}
Our goal is then to allow parties to build a prediction model jointly while protecting all parties from leaking data information to other parties. In contrast with most existing work on privacy-preserving data mining and machine learning, the complexity in our setting is significantly increased. Unlike sample-partitioned/horizontal FL, the vertical FL setting requires a more complex mechanism to decompose the loss function at each party~\mbox{~\cite{vaidya2008survey,vaidya2005privacy,hardy2017private}}. In addition, since only one data provider owns the label information,  we need to propose a secure protocol to guide the learning process instead of sharing label information explicitly among all parties. Finally, data confidentiality and privacy concerns prevent parties from exposing their own users. Hence, entity alignment should also be conducted in a sufficiently secure manner.
\ignore{This would require that the learning system find out the overlapping part of instances across different parties while keeping the non-overlapping part secret.}

\ignore{Different from the above methods which address the problem of privacy protection during the process of user data collection, privacy protection for multiple databases to mine the union of their data is suffering from serious lack of attention. Given the growing reality that collecting user information is crucial to making good software, especially in an age of big data, we can observe "Matthhew effect" in machine learning. The company with more customers is able to collect more data, which is helpful in building a more accurate model. As a result, the company can greatly improve the user experience and attract more users in return, which contributes to richer data. To alleviate the impact of "Matthhew effect", it is critical to encourage the small enterprises to collaborate with each other to mine the union of their data. However, due to the privacy constraint and competition among enterprises, different data providers cannot and will not reveal their information during the learning process. It is necessary for us to propose a solution that solves the general problem of sharing data between different organizations.}

Tree boosting is a highly effective and widely used machine learning method, which excels in many machine learning tasks due to its high efficiency as well as strong interpretability. For example, XGBoost~\cite{xgboost} has been widely used in various applications including credit risk analysis and user behavior studies. In this paper, we propose a novel end-to-end privacy-preserving tree-boosting algorithm and framework known as SecureBoost to enable machine learning in a federated setting. 
Secureboost has been implemented in an open-sourced FL project, FATE\footnote{https://github.com/FederatedAI/FATE} to enable industrial applications.
Our federated learning framework operates in two steps.  First, we find the common users among the parties under a privacy-preserving constraint. Then, we collaboratively learn a shared classification or regression model without leaking any user information to each other. We summarize our main contributions as follows:

\begin{itemize}
\item We formally define a novel problem of privacy-preserving machine learning over vertically partitioned data in the setting of federated learning.
\ignore{\item We proposed a privacy-preserving entity alignment algorithm to ensure zero leakage of non-overlapping part of instances across different parties.}
\item We present an approach to train a high-quality tree boosting model collaboratively while keeping the training data local over multiple parties. Our protocol does not need the participation of a trusted third party.
\item Finally and importantly, we prove that our approach is {\em lossless} in the sense that it is as accurate as any centralized non-privacy-preserving methods that bring all data to a central location.
\item In addition, along with a proof of security, we discuss what would be required to make the protocols completely secure.
\end{itemize}

\section{Preliminaries and Related Work}\label{sec:Related Work}

To protect the privacy of the data used for learning a model, the authors in~\mbox{\cite{shokri2015privacy}} proposed to take advantage of differential privacy (DP) for learning a deep learning model. 
Recently, Google introduced a federated learning framework to prevent the data from being transmitted by bringing the model training to each mobile terminal~\mbox{\cite{konevcny2016federated,Bonawitz:2017:PSA:3133956.3133982,google_workshop_2019}}. Its basic idea is that each local mobile terminal trains the local model using its local data with the same model architecture. The global model can simply be updated by averaging all the local models. Following the same idea, several attempts have been made to reinvent different machine learning models to the federated setting, including decision tree~\mbox{\cite{zhao2018inprivate, li2020practical}}, linear/logistic regression~\mbox{\cite{li2018federated, hanzely2020lower,mohri2019agnostic}} and neural network~\mbox{\cite{yurochkin2019bayesian, wang2020federated}}.


All the above methods are designed for horizontally partitioned data. Unlike sample-partitioned/horizontal FL, the vertical FL setting requires a more complex mechanism to decompose the loss function at each party. The concept of vertical FL is first proposed in~\mbox{ \cite{Yang-et-al:2019,hardy2017private}} and protocols are proposed for linear models~\mbox{\cite{hardy2017private,communication-YangLiu2019}} and neural networks \cite{ftl}. Some previous works have been proposed for privacy-preserving decision trees over vertically partitioned data~\mbox{\cite{vaidya2005privacy,vaidya2008privacy}}. However, their proposed methods have to reveal class distribution over given attributes, which will cause potential security risks. In addition, they can only handle discrete data, which is less practical for real-life scenarios. In contrast, our method guarantees better protection to the data and can be easily applied to continuous data. Another work proposed in~\mbox{\cite{djatmiko2017privacy}} jointly performs logistic regression over the encrypted vertically-partitioned data by approximating a non-linear logistic loss by a Taylor expansion, which will inevitably compromise the performance of the model. In contrast to these works, we propose a novel approach that is \mbox{\em lossless} in nature. 

\section{Problem Statement}\label{sec:Problem Statement}
Let $\left \{ \mathbf{X}^{k} \in \mathbb{R}^{n_{k} \times d_{k}}\right \}_{k=1}^{m}$ be the data matrix distributed on $m$ private parties with each row $\mathbf{X}^{k}_{i*}\in \mathbb{R}^{1 \times d_{k}}$ being a data instance. We use $\mathcal{F}^{k} = \left \{ f_{1},...,f_{d_{k}} \right \}$ to denote the feature set of corresponding data matrix $\mathbf{X}^{k}$. Two parties $p$ and $q$ have different sets of features, denoted as $\mathcal{F}^{p} \cap \mathcal{F}^{q} = \varnothing, \forall p \neq q \in \left \{ 1...m \right \}$. Different parties may hold different sets of users as well, allowing some degree of overlap. Only one of the parties holds the class labels $\mathbf{y}$.

\begin{definition}\textbf{Active Party:}\end{definition} We define the active party as the data provider who holds both a data matrix and the class label. Since the class label information is indispensable for supervised learning, the active party naturally takes the responsibility as a dominating {\em server} in federated learning.

\begin{definition}\textbf{Passive Party:}\end{definition} We define the data provider which has only the data matrix as a passive party. Passive parties play the role of clients in the federated learning setting.  

The problem of privacy-preserving machine learning over vertically split data in federated learning can be stated as:

\textbf{Given:} a vertically partitioned data matrix $\left \{ \mathbf{X}^{k} \right \}_{k=1}^{m}$ distributed on $m$ private parties and the class labels $\mathbf{y}$ distributed on active party.

\textbf{Learn:} a machine learning model $M$ without giving information of the data matrix of any party to others in the process.  The model $M$ is a function that has a projection $M_i$ at each party $i$, such that $M_i$ takes input of its own features $X_i$.

\textbf{Lossless Constraint:} We require that the model $M$ is lossless, which means that the loss of $M$ under federated learning over the training data is the same as the loss of $M'$ when $M'$ is built on the union of all data.

\section{Federated Learning with SecureBoost}\label{sec:Proposed Approach: SecureBoost}
As one of the most popular machine learning algorithms, the gradient-tree boosting excels in many machine learning tasks, such as fraud detection, feature selection and product recommendation. In this section, we propose a novel gradient-tree boosting algorithm called SecureBoost in the federated learning setting. It consists of two major steps.  First, it aligns the data under the privacy constraint.  Second, it collaboratively learns a shared gradient-tree boosting model while keeping all the training data secret over multiple private parties.  We explain each step below.

Our first goal is to find a common set of data samples at all participating parties so as to build a joint model $M$.  When the data is vertically partitioned among parties, different parties hold different but partially overlapping users, which can be identified using their IDs. The problem is how to find the common data samples across the parties without revealing the non-shared parts. To achieve this goal, we align the data samples under a privacy-preserving protocol for inter-database intersections~\cite{liang2004privacy}.

After aligning the data across different parties under the privacy constraint, we now consider the problem of jointly building tree ensemble models over multiple parties without violating privacy. Before further discussing the detail of the algorithm, we first introduce the general framework of federated learning. In federated learning, a typical iteration consists of four steps. First, each client downloads the current global model from server. Second, each client computes an updated model based on its local data and the current global model, which resides within the active party. Third, each client sends the model update back to the server under encryption. Finally, the server aggregates these model updates and construct the updated global model.

Following the general framework of federated learning, we see that to design a privacy-preserving tree boosting framework in the setting of federated learning, essentially we have to answer the following three questions: (1) How can each client (i.e., a passive party) compute an updated model based on its local data without reference to class label? (2) How can the server (i.e., the active party) aggregate all the updated model and obtain a new global model? (3) How to share the updated global model among all parties without leaking any information at inference time? To answer these three questions, we start by reviewing a tree ensemble model, XGBoost~\cite{chen2016xgboost}, in a non-federated setting.

Given a data set $\mathbf{X}\in\mathbb{R}^{n\times d}$ with $n$ samples and $d$ features, XGBoost predicts the output by using $K$ regression trees.
\begin{equation}\label{eq:General Tree Boosting Framework}
\hat{y_{i}} = \sum_{k=1}^{K}f_{k}(\mathbf{x}_i)
\end{equation}

To learn the set of regression tree models used in Eq.(\ref{eq:General Tree Boosting Framework}), it greedily adds a tree $f_{t}$ at the $t$-th iteration to minimize the following loss.
\begin{equation}\label{eq:Xgboost}
\mathcal{L}^{(t)}\simeq \sum_{i=1}^{n}\left[l\left(y_{i},\hat{y_{i}}^{(t-1)}\right)+g_{i}f_{t}\left(\mathbf{x}_{i}\right)+\frac{1}{2}h_{i}f_{t}^{2}\left(\mathbf{x}_{i}\right)\right]+\Omega (f_{t})
\end{equation}
where $\Omega (f_{t}) = \gamma T+\frac{1}{2}\lambda \left \| w \right \|^{2}$, $g_{i} = \partial_{\hat{y}^{(t-1)}} l(y_{i},\hat{y}^{(t-1)})$ and $h_{i} = \partial_{\hat{y}^{(t-1)}}^{2} l(y_{i},\hat{y}^{(t-1)})$.

When constructing the regression tree in the $t$-th iteration, it starts from the tree with depth of $0$ and add a split for each leaf node until reaching the maximum depth. In particular, it maximizes the following equation to determine the best split, where $I_{L}$ and $I_{R}$ are the instance spaces of left and right tree nodes after the split. 

\begin{equation}\label{eq:Split Finding}
\mathcal{L}_{sp} = \frac{1}{2}\left[\frac{\left(\sum_{i \in I_{L}}g_{i}\right)^2 }{\sum_{i \in I_{L}}h_{i}+\lambda}+\frac{\left(\sum_{i \in I_{R}}g_{i}\right)^2 }{\sum_{i \in I_{R}}h_{i}+\lambda}-\frac{\left(\sum_{i \in I}g_{i}\right)^2 }{\sum_{i \in I}h_{i}+\lambda  }\right]-\gamma
\end{equation}

After it obtains an optimal tree structure, the optimal weight $w_{j}^{*}$ of leaf $j$ can be computed by the following equation, where $I_{j}$ is the instance space of leaf $j$.
\begin{equation}\label{eq:w}
w_{j}^{*}=-\frac{\sum_{i\in I_{j}}g_{i}}{\sum_{i\in I_{j}}h_{i}+\lambda}
\end{equation}

From the above review, we make following observations:

(1) The evaluation of split candidates and the calculation of the optimal weight of leaf only depends on the $g_{i}$ and $h_{i}$, which makes it easily adapted to the setting of federated learning. 

(2) The class label can be inferred from $g_{i}$ and $h_{i}$. For instance, when we take the square loss as the loss function, we have $g_{i} = \hat{y}_i^{(t-1)} - y_i$.

With the above observations, we now introduce our federated gradient tree boosting algorithm. Following observation (1), we can see that passive parties can determine their locally optimal split with only its local data and $g_{i}$,$h_{i}$, which motivates us to follow such method to decompose learning task at each party. However, according to observation (2), $g_{i}$ and $h_{i}$ should be regarded as sensitive data, since they are able to disclose class label information to passive parties. Therefore, in order to keep $g_{i}$ and $h_{i}$ confidential, the active party is required to encrypt $g_{i}$ and $h_{i}$ before sending them to passive parties. The remaining challenge is how to determine the locally optimal split with encrypted $g_{i}$ and $h_{i}$ for each passive party.

According to Eq.(\ref{eq:Split Finding}), the optimal split can be found if $g_{l} = \sum_{i \in I_{L}}g_{i}$ and $h_{l} = \sum_{i \in I_{L}}h_{i}$ are  calculated for every possible split. So next, we show how to obtain $g_{l}$ and $h_{l}$ with encrypted $g_{i}$ and $h_{i}$ using additive homomorphic encryption scheme~\cite{paillier1999public}. The Paillier encryption scheme is taken as our encryption scheme.

\begin{algorithm}[!t]
    \caption{Aggregate Encrypted Gradient Statistics}
    \label{alg:Aggregate Secure Gradient Statistics}
    \begin{algorithmic}[1]
        \Require $I$, instance space of current node
        \Require $d$, feature dimension
        \Require $\left \{ \left \langle g_{i} \right \rangle ,\left \langle h_{i} \right \rangle \right \}_{i \in I}$
        \Ensure $\mathbf{G} \in \mathbb{R}^{d \times l}$, $\mathbf{H} \in \mathbb{R}^{d \times l}$

        \For{$k = 0 \to d$}
            \State Propose $S_{k} = \left \{s_{k1},s_{k2},...,s_{kl}\right \}$ by percentiles on feature $k$
        \EndFor

        \For{$k = 0 \to d$}
            \State $\mathbf{G}_{kv} = \sum_{i \in \left \{ i|s_{k,v}\geq x_{i,k}> s_{k,v-1}\right \}}\left \langle g_{i} \right \rangle$
            \State $\mathbf{H}_{kv} = \sum_{i \in \left \{ i|s_{k,v}\geq x_{i,k}> s_{k,v-1}\right \}}\left \langle h_{i} \right \rangle$
        \EndFor

    \end{algorithmic}
\end{algorithm}

Denoting the encryption of a number $u$ under the Paillier cryptosystem as $\left \langle u \right \rangle$, the main property of the Paillier cryptosystem ensures that for arbitrary numbers $u$ and $v$, we have $\left \langle u \right \rangle . \left \langle v \right \rangle = \left \langle u+v \right \rangle$. Therefore, $\left \langle h_{l} \right \rangle = \prod_{i \in I_{L}}\left \langle h_{i} \right \rangle$ and $\left \langle g_{l} \right \rangle = \prod_{i \in I_{L}}\left \langle g_{i} \right \rangle$. Consequently, the best split can be found in the following way. First, each passive party computes $\left \langle g_{l} \right \rangle$ and $\left \langle h_{l} \right \rangle$ for all possible splits locally, which are then sent back to the active party. The active party deciphers all $\left \langle g_{l} \right \rangle$ and $\left \langle h_{l} \right \rangle$ and calculates the global optimal split according to Eq.(\ref{eq:Split Finding}). We adopt the approximation scheme used by~\cite{chen2016xgboost}, so as to alleviate the need of enumerating all possible split candidates and communicating their $\langle g_i\rangle$ and $\langle h_i \rangle$. The details of our secure gradient aggregation algorithm are shown in Algorithm~\ref{alg:Aggregate Secure Gradient Statistics}.

Following the observation (1), the split finding algorithm remains largely the same as XGBoost except for minor adjustments to fit the federated learning framework. 
Due to separation in features, SecureBoost requires different parties to store certain information for each split, so as to perform prediction for new samples. Passive parties should keep a lookup table as shown in Figure~{\ref{Fig.SecureBoostTree}}. It contains split thresholds [feature id $k$, threshold value $v$] and a unique record id $r$ used to index the table, in order to look up split conditions during inference. In the meantime, because the active party does not have features located in passive parties, for the active party to know which passive party to deliver an instance to, as well as instructing the passive party which split condition to use at inference time, it associates every tree node with a pair (party id $i$, record id $r$). Specific details about the split finding algorithm for SecureBoost is summarized in Algorithm~{\ref{alg:Split Finding}}. The problem remaining is the computation of optimal leaf weights. According to Equation~{\ref{eq:w}}, the optimal weight of leaf $j$ only depends on $\sum_{i\in I_j}g_i$ and $\sum_{i\in I_j}h_i$. Consequently, it follows similar procedures as split finding. When a leaf node is reached, the passive party sends $\langle \sum_{i\in I_j}g_i\rangle$ and $\langle\sum_{i\in I_j}h_i\rangle$ to the active party, which are then deciphered to compute corresponding weights through Equation~{\ref{eq:w}}.

\ignore{After the active party obtains the global optimal split, [party id ($i$), feature id ($k$), threshold id ($v$)], it returns $k$ and $v$ to the corresponding passive party $i$. Passive party $i$ decides the selected attribute's value based on $k$ and $v$, and then partitions the current instance space according to the selected attribute's value. In addition, it builds a lookup table locally to record the selected attribute's value, [feature $k$, threshold value $v$], as shown in Figure~\ref{Fig.SecureBoostTree}. After that, it returns the index of the record and the split instance space (i.e. $I_{L}$, $I_R$) back to the active party. The active party splits the current node according to the received instance space and associate current node with [party id, record id], until a stopping criterion or the max depth is reached. \ignore{All the leaf nodes are stored inside the active party.} The detailed algorithm for our secure federated tree boosting framework is summarized in Algorithm~\ref{alg:Split Finding}.}

\begin{algorithm}[!t]
    \caption{Split Finding}
    \label{alg:Split Finding}
    \begin{algorithmic}[1]

        \Require I, instance space of current node
        \Require $\left \{ \mathbf{G}^{i},\mathbf{H}^{i}\right \}_{i=1}^{m}$, aggregated encrypted gradient statistics from $m$ parties
        \Ensure Partition current instance space according to the selected attribute's value\\
        /*Conduct on Active Party*/\\
        $g \gets \sum_{i \in I} g_{i},h \gets \sum_{i \in I} h_{i}$
        \For{$i = 0$ to $m$}
            \For{$k = 0$ to $d_{i}$}
                \State $g_{l} \gets 0,h_{l} \gets 0$

                \State //enumerate all threshold value

                \For{$v = 0$ to $l_{k}$}
                    \State get decrypted values $D(\mathbf{G}_{kv}^{i})$ and $D(\mathbf{H}_{kv}^{i})$
                    \State $g_{l} \gets g_{l}+D(\mathbf{G}_{kv}^{i}),h_{l} \gets h_{l}+D(\mathbf{H}_{kv}^{i})$
                    \State $g_{r} \gets g- g_{l},h_{r} \gets h- h_{l}  $
                    \State $score \gets \max(score, \frac{g_{l}^{2}}{h_{l}+\lambda}+\frac{g_{r}^{2}}{h_{r}+\lambda}-\frac{g^{2}}{h+\lambda})$
                \EndFor
            \EndFor
        \EndFor\\

Return $k_{opt}$ and $v_{opt}$ to the passive party $i_{opt}$ when we obtain the max score.\\

/*Conduct on Passive Party $i_{opt}$*/\\

Determine the selected attribute's value according to $k_{opt}$ and $v_{opt}$ and partition current instance space.\\

Record the selected attribute's value and return [record id, $I_{L}$] back to the active party.\\

/*Conduct on Active Party*/\\

Split current node according to $I_{L}$ and associate current node with [party id, record id].

\end{algorithmic}
\end{algorithm}

\section{Federated Inference}\label{sec:Federated Inference}
\begin{figure*}[!t]
\centering
\includegraphics[width=0.8\linewidth]{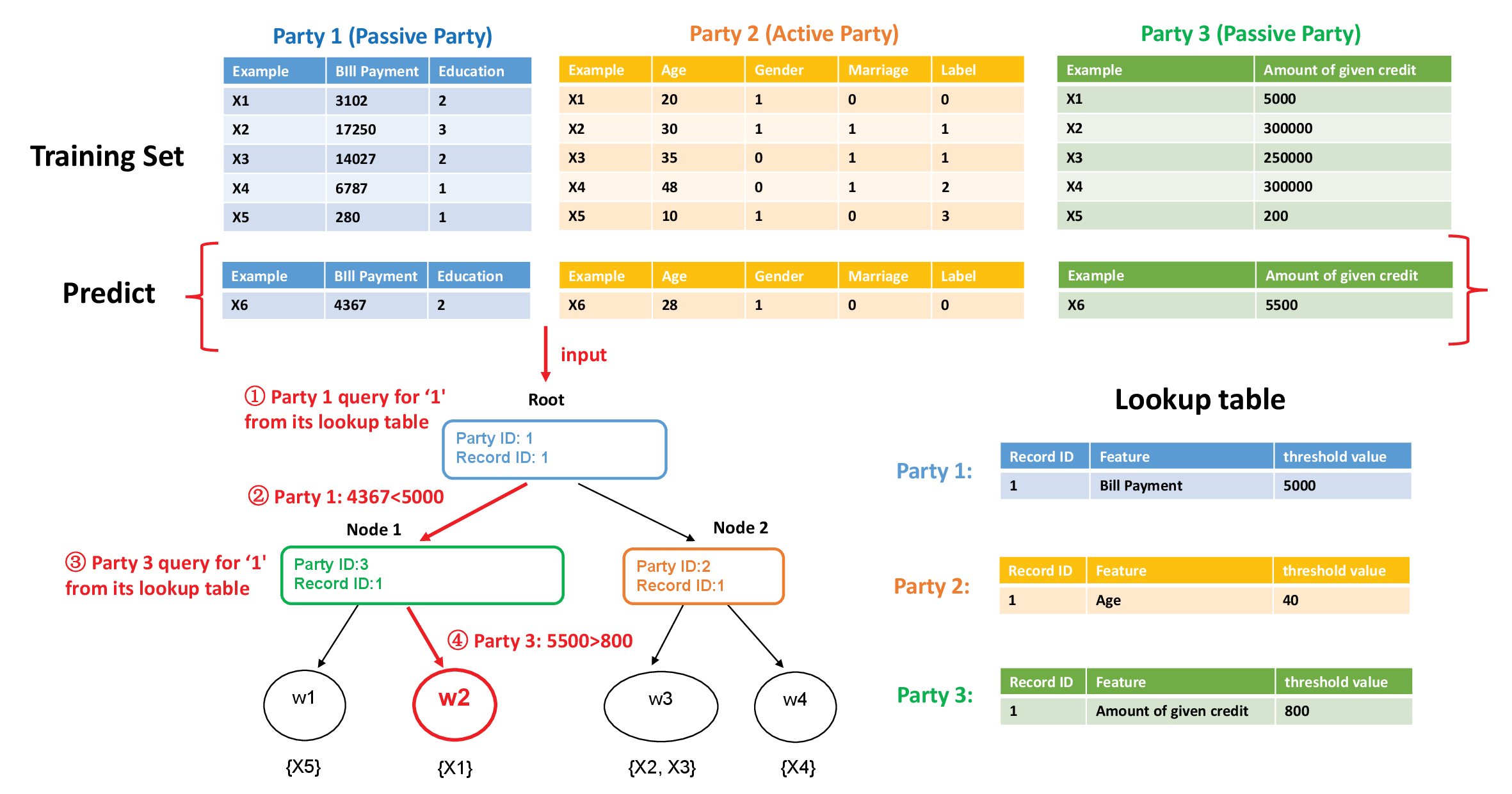}
\caption{An illustration of Federated Inference}\label{Fig.SecureBoostTree}
\end{figure*}
In this section, we describe how the learned model (distributed among parties) can be used to classify a new instance even though the features of the instance to be classified are private and distributed among parties. Since each party knows its own features but nothing of the others, we need a secure distributed inference protocol to control passes from one party to another, based on the decision made. To illustrate the inference process, we consider a system with three parties as depicted in Figure~\ref{Fig.SecureBoostTree}. Specifically, party $1$ is the active party, which collects information including user's \textit{monthly bill payment, education}, as well as the \textit{label}, whether the user $X$ made the payment on time. Party $2$ and party $3$ are passive parties, holding features \textit{age, gender, marriage status} and \textit{amount of given credit} respectively. Suppose we wish to predict whether a user $X_6$ would make payment on time, then all sites would have to collaborate to make the prediction. The whole process is coordinated by the active party. Starting from the root, by referring to the record [party id:$1$, record id:$1$], the active party knows party $1$ holds the root node, thereby requiring party $1$ to retrieve the corresponding attribute, \textit{Bill Payment}, from its lookup table based on the record id $1$. Since the classifying attribute is \textit{bill payment} and party $1$ knows the bill payment for user $X6$ is $4367$, which is less than the threshold $5000$, it makes the decision that it should move down to its left child, node $1$. Then, active party refers to the record [party id:3, record id:1] associated with node $1$ and requires party $3$ to conduct the same operations. This process continues until a leaf is reached.


\section{Theoretical Analysis for Lossless Property}
\begin{theorem}\label{theorem:1}
SecureBoost is lossless, i.e. SecureBoost model $M$ and XGBoost model $M'$ would behave identically provided that the models $M$ and $M'$ are identically initialized and hyper-parameterized.
\end{theorem}

\begin{proof}
According to Eq.(\ref{eq:Split Finding}), $g_{l}$ and $h_{l}$ are the only information needed for the calculation of the best split, which can be obtained with encrypted $g_{i}$ and $h_{i}$ using Paillier cryptosystem in SecureBoost. In the Paillier cryptosystem, the encryption of a message $m$ is $\left \langle m \right \rangle=g^{m}r^{n} \mod n^{2}$, for some random $r\in \{0,\ldots ,n-1\}$. Given the definition of encrypted message, we have $\left \langle m_1 \right \rangle . \left \langle m_2 \right \rangle = \left \langle m_1+m_2 \right \rangle$ for arbitrary message $m_1$ and $m_2$ under Paillier cryptosystem, which can be proved as follows:

\begin{equation}
\begin{aligned}
    \left \langle m_1 \right \rangle . \left \langle m_2 \right \rangle&=(g^{m_1}r^{c}_{1})(g^{m_2}r^{c}_{2})\mod n \\
    &= g^{m_1+m_2}(r_1r_2)^c \mod n \\
    &= \left \langle m_1+m_2 \right \rangle
\end{aligned}
\end{equation}

Therefore, we have $\left \langle h_{l} \right \rangle = \prod_{i \in I_{L}}\left \langle h_{i} \right \rangle$ and $\left \langle g_{l} \right \rangle = \prod_{i \in I_{L}}\left \langle g_{i} \right \rangle$. Provided that with the same initialization, an instance $i$ will have the same value of $g_{i}$ and $h_{i}$ under either setting. Thus, model $M$ and $M'$ can always achieve the same best split throughout the construction of the tree and result in identical $M$ and $M'$, which ensures the property of lossless.
\end{proof}

\section{Security Discussion}\label{sec:Security Discussion}
SecureBoost avoids revealing data records held by each of the parties to others during training and inference thus protecting the privacy of individual parties' data. However, we stress that there is some leakage that can be inferred during the protocol execution which is quite different for passive vs. active parties. 
\ignore{Due to space limitations, we analyze this leakage in more detail in the extended version of the paper, and describe it here at a high level.}

The \textbf{active party} is in an advantageous position with SecureBoost as it learns the instance space for each split and which party is responsible for the decision at each node. Also, it learns all the possible values of $g_{l},g_r$ and $h_{l},h_r,$ during learning. The former seems unavoidable in this setting, unless one is willing to severely increase the overhead during the inference phase. However, the latter can be avoided using secure multi-party computation techniques for comparison of encrypted values (e.g.,~\cite{bostPTG15,baldimtsiPPST17}). In this way, the active party learns only the optimal $g_{l},g_r,h_{l},h_r$ per party; on the other hand, this significantly affects the efficiency during learning.

Note that the instances that are associated with the same leaf strongly indicates they belong to the same class. We denote the proportion of samples which belong to the majority class as \textbf{leaf purity}. The information leakage with respect to \textbf{passive parties} is directly related with \textbf{leaf purity} of the first tree of SecureBoost. Moreover, the  first tree's \textbf{leaf purity} can be inferred from the weight of the leaves.

\ignore{Hence, in order to reduce information leakage with respect to \textbf{passive parties}, we opt to store decision tree leaves at the active party. However, this does not entirely eliminate this leakage (e.g., consider a passive party holds the parent node of two leaf nodes). \ignore{In the extended version of our paper,} We prove that such leakage is directly related with leaf purity of the produced tree (this refers to the proportion of samples which belong to the majority class), which can be inferred from the weight of the first tree's leaves.}


\begin{theorem}\label{theorem:2}
Given a learned SecureBoost model, its first tree's leaf purity can be inferred from the weight of the leaves.
\end{theorem}

\begin{proof}
The loss function for binary classification problem is given as follows.

\begin{equation}
L = y_{i}log(1+e^{-\hat{y_{i}}})+(1-y_{i})log(1+e^{\hat{y_{i}}})
\end{equation}

Based on the loss function, we have $g_{i} = \hat{y_{i}}^{(0)}-y_{i}$ and $h_{i} = \hat{y_{i}}^{(0)}*(1-\hat{y_{i}}^{(0)})$ during the construction of the decision tree at first iteration. Specifically, $\hat{y_{i}}^{(0)}$ is given as initialized value. Suppose we initialize all $\hat{y_{i}}^{(0)}$ as $a$ where $0<a<1$. According to Eq.(\ref{eq:w}), for the instances associated with the specific leaf $j$, $\hat{y_{i}}^{(1)} = S(w_{j}^{*})=S(-\frac{\sum_{i\in I_{j}}g_{i}}{\sum_{i\in I_{j}}h_{i}+\lambda})$ where $S(x)$ is the sigmoid function. Suppose the number of instances associated with the leaf $j$ is $n_{j}$ and the percentage of positive samples is $\theta_{j}$. When $n_{j}$ is relatively big, we can ignore $\lambda$ in $-\frac{\sum_{i\in I_{j}}g_{i}}{\sum_{i\in I_{j}}h_{i}+\lambda}$ and rewrite the weight of leaf $j$ as $w_{j}^{*}=-\frac{\sum_{i\in I_{j}}g_{i}}{\sum_{i\in I_{j}}h_{i}} = -\frac{\theta_j*n*(a-1)+(1-\theta_j)*n*a}{n*a*(1-a)} = -\frac{\theta_j*n*(a-1)+(1-\theta_j)*n*a}{n*a*(1-a)} = \frac{a-\theta_j}{a(a-1)}$. By reformulating the equation, we have $\theta_j = a-a(a-1)w_{j}^{*}$. $\theta_j$ depends on $a$ and $w_{j}^{*}$ and $a$ is given as initialization. Thus, $w_{j}^{*}$ is the key to determine $\theta_j$. Note that $\theta_j$ can be used to represent the leaf purify of leaf $j$ (i.e., purify of leaf $j$ can be formally written as $\max(\theta_j, 1-\theta_j)$, leaf purity of the first tree can be inferred from the weight of the leaves ($w_{j}^{*}$) given a learned SecureBoost model.
\end{proof}

\ignore{According to Theorem~\ref{theorem:2}, as long as weight of the first tree's leaves are close enough to $S(\frac{2a-1}{2a(a-1)})$, the protocol is considered secure.}


According to Theorem~\ref{theorem:2}, given a SecureBoost model, the weight of the leaves of its first tree can reveal sensitive information. In order to reduce information leakage with respect to \textbf{passive parties}, we opt to store decision tree leaves at the active party and propose a modified version of our framework, called \textit{Reduced-Leakage SecureBoost (RL-SecureBoost)}. With \textit{RL-SecureBoost}, the active party learns the first tree independently based only on its own features which fully protects the instance space of its leaves. Hence, all the information that passive parties learn is based on residuals. Although the residuals may also reveal information, we prove that as the purity in the first tree increases, this residual information decreased. 

\begin{theorem}\label{theorem:3}
As the purity in the first tree increases, the residual information decreased. 
\end{theorem}

\begin{proof}
As mentioned before, for binary classification problem, we have $g_{i} = \hat{y_{i}}^{(t-1)}-y_{i}$ and $h_{i} = \hat{y_{i}}^{(t-1)}*(1-\hat{y_{i}}^{(t-1)})$, where $g_{i} \in [-1,1]$. Hence,

\begin{equation}
\left\{\begin{matrix}
h_{i} = g_{i}(1-g_{i}), & \text{if } y_{i}=0 \\ 
h_{i} = -g_{i}(g_{i}+1), & \text{if } y_{i}=1
\end{matrix}\right.
\end{equation}

When we construct the decision tree at the t-th iteration with $k$ leaves to fit the residuals of the previous tree, in essential, we split the data into $k$ clusters to minimize the following loss.

\begin{equation}
\begin{split}
L =& -\sum _{j=1}^{k}\frac{(\sum_{i\in I_{j}}g_{i})^2}{\sum_{i\in I_{j}}h_{i}}\\
=& -\sum _{j=1}^{k}\frac{(\sum_{i\in I_{j}}g_{i})^2}{\sum_{i\in I_{j}^{N}} g_{i}(1-g_{i})+\sum_{i\in I_{j}^{P}} -g_{i}(1+g_{i})}\\
\end{split}
\end{equation}

We know $\hat{y_{i}}^{(t-1)} \in [0,1]$ and $g_{i} = \hat{y_{i}}^{(t-1)}-y_{i}$. Thus, we have $g_{i}\in [-1,0]$ for positive samples and $g_{i}\in [0,1]$ for negative samples. Taking the range of $g_{i}$ into consideration, we can rewrite the above equation as follows.

\begin{equation}\label{eq:residuals_1}
\sum _{j=1}^{k}\frac{(\sum_{i\in I_{j}^{N}}|g_{i}|-\sum_{i\in I_{j}^{P}} |g_{i}|)^2}{\sum_{i\in I_{j}^{N}} |g_{i}|(|g_{i}|-1)+\sum_{i\in I_{j}^{P}}|g_{i}|(|g_{i}|-1)}
\end{equation}

Where $I_{j}^{N}$ and $I_{j}^{P}$ denote the set of negative samples and positive samples associated with leaf $j$ respectively. We denote the expectation of $|g_{i}|$ for positive samples as $\mu_{p}$ and the expectation of $|g_{i}|$ for negative samples as $\mu_{n}$. When we have a large amount of samples but small number of leave nodes $k$, we can use the following equation to approximates Eq.(~\ref{eq:residuals_1}).

\begin{equation}\label{eq:residuals_2}
\sum _{j=1}^{k}\frac{(n_{j}^{n}\mu_{n}-n_{j}^{p} \mu_{p})^2}{ n_{j}^{n}\mu_{n}(\mu_{n}-1)+n_{j}^{p}\mu_{p}(\mu_{p}-1)}
\end{equation}

Where $n_{j}^{n}$ and $n_{j}^{p}$ represent the number of negative samples and positive samples associated with leaf $j$. Since $\mu_{n}\in [0,1]$ and $\mu_{n}\in [0,1]$, we know the numerator has to be positive and the denominator has to be negative. Thus, the whole equation has to be negative. To minimize Eq.(\ref{eq:residuals_2}) is equal to maximizing the numerator while minimizing the denominator. Note that the denominator is $\sum x^2$ and the numerator is $(\sum x)^2$ where $x \in [0,1]$ . The equation is dominated by numerator. Thereby, minimizing Eq.(~\ref{eq:residuals_2}) can be regarded as maximizing the numerator $(n_{j}^{n}\mu_{n}-n_{j}^{p} \mu_{p})^2$. Ideally, we require $n_{j}^{n} = n_{j}^{p}$ in order to prevent label information from divulging. When $|\mu_{n}-\mu_{p}|$ is bigger, more possible we can achieve the goal. And we know $|g_{i}|=|\hat{y_{i}}^{(t-1)}-y_{i}| = \hat{y_{i}}^{(t-1)}$ for negative samples and $|g_{i}|=|\hat{y_{i}}^{(t-1)}-y_{i}| = 1-\hat{y_{i}}^{(t-1)}$ for positive samples. Thereby, $\mu_{n} = \frac{1}{N_n}\sum_{j=1}^{k}(1-\theta_{j})n_{j}\hat{y_{i}}^{(t-1)}$ and $\mu_{p} = \frac{1}{N_p}\sum_{j=1}^{k}\theta_{j}n_{j}(1-\hat{y_{i}}^{(t-1)})$. $|\mu_{n}-\mu_{p}|$ can be calculated as follows.

\begin{equation}\label{eq:residuals_3}
\begin{split}
&|\mu_{n}-\mu_{p}| \\ =&|\frac{1}{N_n}\sum_{j=1}^{k}(1-\theta_{j})n_{j}\hat{y_{i}}^{(t-1)}-\frac{1}{N_p}\sum_{j=1}^{k}\theta_{j}n_{j}(1-\hat{y_{i}}^{(t-1)})|
\end{split}
\end{equation}

Where $N_n$ and $N_p$ correspond to the number of negative samples and positive samples in total. $\theta_{j}$ is the percentage of positive samples associated with leave $j$ for decision tree at $(t-1)$-th iteration (previous decision tree). $n_{j}$ denote the number of instances associated with leave $j$ for previous decision tree. $\hat{y_{i}}^{(t-1)} = S(w_{j})$ where $w_{j}$ represents the weight of $j$-th leave of previous decision tree. When the positive samples and negative samples are balanced, $N_n = N_p$, we have

\begin{equation}\label{eq:residuals_4}
\begin{split}
&|\mu_{n}-\mu_{p}| \\
=&\frac{1}{N_n}|\sum_{j=1}^{k}((1-\theta_{j})n_{j}S(w_{j})-\theta_{j}n_{j}(1-S(w_{j}))| \\
=&\frac{1}{N_n}\sum_{j=1}^{k}n_{j}|(S(w_{j})-\theta_{j})|\\
=&\frac{1}{N_n}\sum_{j=1}^{k}n_{j}|(S(\frac{a-\theta_{j}}{a(a-1)})-\theta_{j})|
\end{split}
\end{equation}

As observed from Eq.(~\ref{eq:residuals_4}), it achieves the minimum value when $S(\frac{a-\theta_{j}}{a(a-1)}) = a$. By solving the equation, we have the optimal solution of $\theta_{j}$ as $\theta_{j}* = a(1+(1-a)\ln(\frac{a}{1-a})))$. In order to achieve bigger $\mu_{n}-\mu_{p}$, we want the deviation from $\theta_{j}$ to $\theta_{j}*$ to be as big as possible. When we have proper initialization of $a$, for instance $a = 0.5$, $\theta_{j}* = 0.5$. In this case, maximizing $|\theta_{j}-\theta_{j}*|$ is the same as maximizing $\max(\theta_{j},1-\theta_{j})$, which exactly is the leaf purity. Therefore, we have proved that high leaf purity will guarantee big difference between $\mu_{n}$ and $\mu_{p}$, which finally results in less information leakage. We complete our proof.
\end{proof}

Given Theorem~\ref{theorem:3}, we prove that \textit{RL-SecureBoost} is secure as long as its first tree learns enough information to mask the actual label with residuals. Moreover, as we experimentally demonstrate in Section~\ref{sec:Experiments}, \textit{RL-SecureBoost} performs identically as \textit{SecureBoost} in terms of prediction accuracy.

\section{Experiments}\label{sec:Experiments}
\begin{figure}[!t]
\begin{minipage}{0.48\linewidth}
\centerline{\includegraphics[width=115px,height=2.5cm]{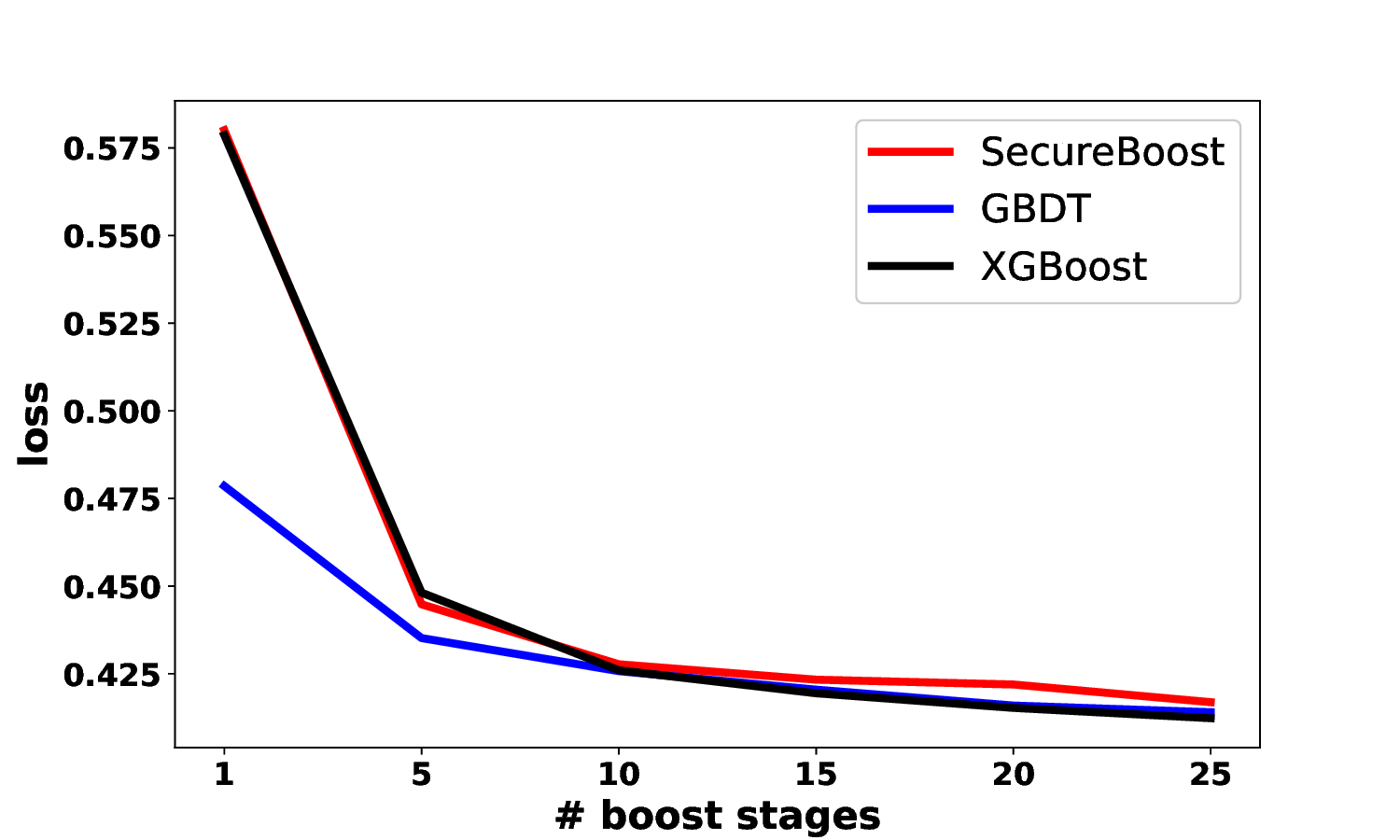}}
\centering{(a) Learning Curve}
\end{minipage}
\hfill
\begin{minipage}{0.48\linewidth}
\centerline{\includegraphics[width=115px,height=2.5cm]{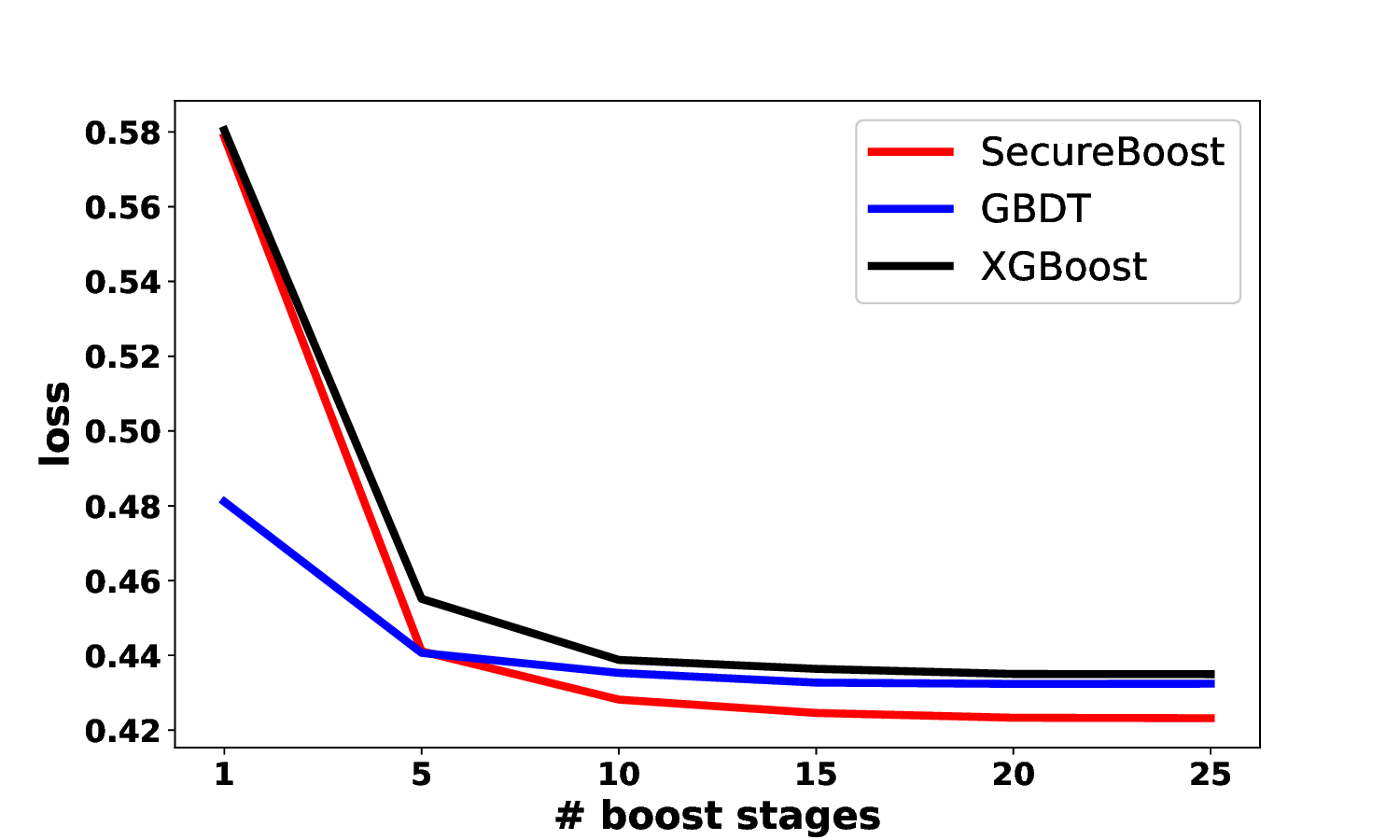}}
\centering{(b) Test Error}
\end{minipage}
\caption{Loss convergence}
\label{Fig:Loss convergence}
\end{figure}

\begin{figure*}
\begin{minipage}{0.28\linewidth}
\centerline{\includegraphics[width=160px,height=3.25cm]{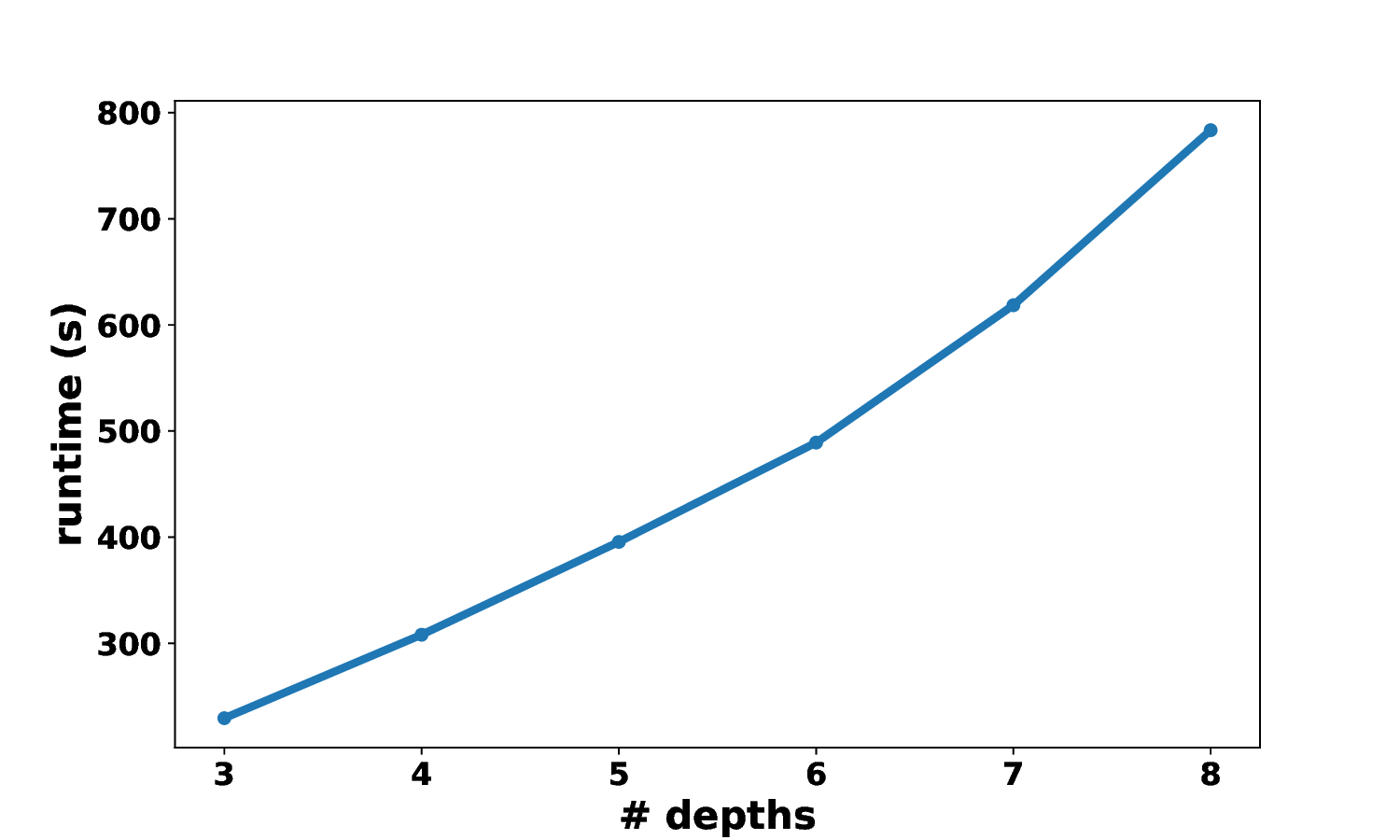}}
\centering{(a)Runtime w.r.t. maximum depth of individual tree}
\end{minipage}
\hfill
\begin{minipage}{0.28\linewidth}
\centerline{\includegraphics[width=160px,height=3.25cm]{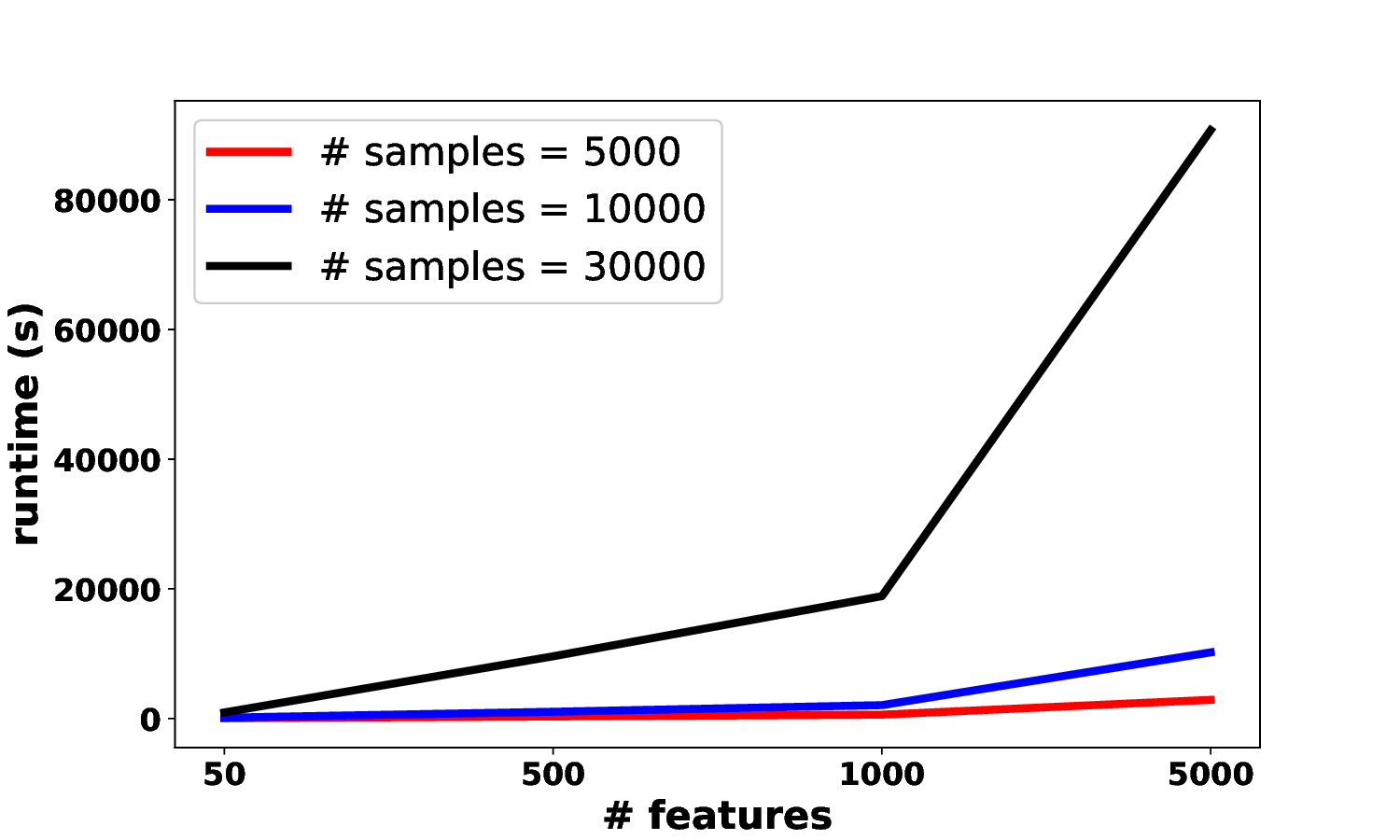}}
\centering{(b)Runtime w.r.t. feature size}
\end{minipage}
\hfill
\begin{minipage}{0.28\linewidth}
\centerline{\includegraphics[width=160px,height=3.25cm]{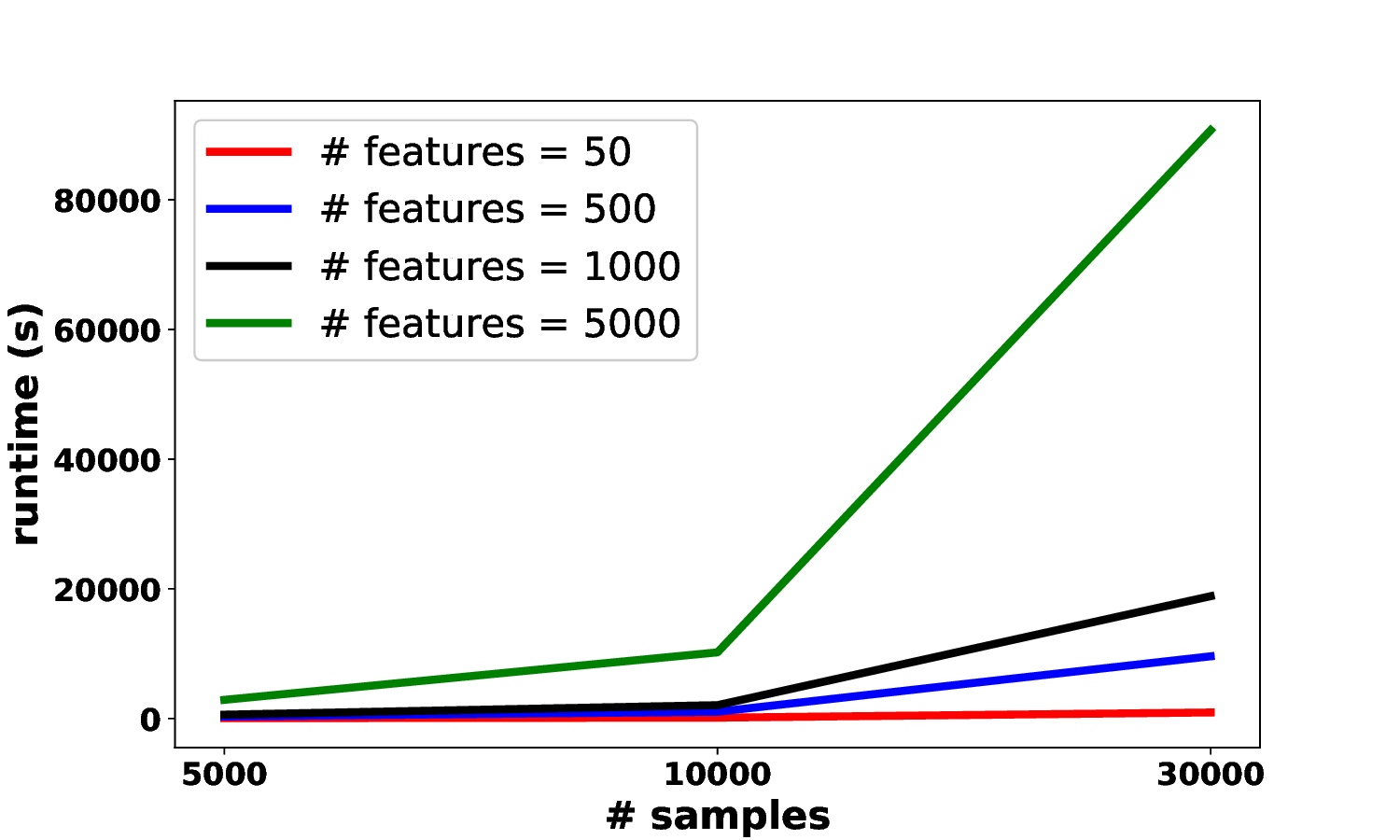}}
\centering{(c)Runtime w.r.t. sample size}
\end{minipage}
\caption{Scalability Analysis of SecureBoost}
\label{Fig:Scalability Analysis}
\end{figure*}

\begin{table}
\centering
\caption{First Tree vs. Second Tree in terms of Leaf Purity}\label{table:Security}
\begin{tabular}{lll} \hline
Mean Purity & Credit 1 & Credit 2 \\ \hline
1st Tree& 0.8058  & 0.7159  \\
2rd Tree& 0.66663 & 0.638 \\ \hline
\end{tabular}
\end{table}

\begin{table*}
\centering\caption{Classification Performance for RL-SecureBoost vs. SecureBoost}\label{table:Accuracy}

    \begin{tabular}{lcccccc}

    \toprule
    \multirow{2}{*}{Model} & \multicolumn{3}{c}{Credit 1} & \multicolumn{3}{c}{Credit 2} \\ 
    \cline{2-7}
    & ACC & F1-score & AUC & ACC & F1-score & AUC \\
    \midrule
    
    1st Tree, SecureBoost & 0.9298 & 0.012 & 0.7002 & 0.7806  & 0 & 0.6381 \\
    1st Tree, RL-SecureBoost & 0.9186 & 0 & 0.6912  & 0.7793  & 0 & 0.6320  \\
    
    \midrule
    
    Overall, SecureBoost & 0.9345  & 0.2576  &  0.8461  & 0.8180 & 0.4634  & 0.7701  \\
    Overall, RL-SecureBoost & 0.9331 & 0.2549 & 0.8423 & 0.8179  & 0.4650  & 0.7682  \\

    \bottomrule 

    \end{tabular}
\end{table*}

We conduct experiments on two public datasets. 

\textbf{Credit $1$}\footnote{https://www.kaggle.com/c/GiveMeSomeCredit/data}:
It involves the problem of classifying whether a user would suffer from serious financial problems. It contains a total of $150000$ instances and $10$ attributes.

\textbf{Credit $2$}\footnote{https://www.kaggle.com/uciml/default-of-credit-card-clients-dataset}:
It is also a credit scoring dataset, correlated to the task of predicting whether a user would make payment on time. It consist of $30000$ instances and $25$ attributes in all.

In our experiment, we use $2/3$ of each dataset for training and the remaining for testing. We split the data vertically into two halves and distribute them to two parties. To fairly compare different methods, we set the maximum depth of each tree as $3$, the fraction of samples used to fit individual regression trees as $0.8$, and learning rate as $0.3$ for all methods. The Paillier encryption scheme is taken as our encryption scheme with a key size of $512$ bits. All experiments are conducted on a machine with $8$GB RAM and Intel Core i5-7200u CPU.

\subsection{Scalability}\label{sec:Scalability}
Note that the efficiency of SecureBoost may be reflected by rate of convergence and runtime, which may be influenced by (1) maximum depth of individual regression trees; (2) the size of the datasets. In this subsection, we conduct convergence analysis as well as study the impact of all the variables on the runtime of learning. All experiments are conducted on dataset Credit$2$.

First, we are interested in the convergence rate of our proposed system. We compare the convergence behavior of SecureBoost with non-federated tree boosting counterparts, including GBDT\footnote{\url{http://scikit-learn.org/stable/modules/generated/sklearn.ensemble.GradientBoostingClassifier.html}} and XGBoost\footnote{https://github.com/dmlc/xgboost}.  As can be observed from Figure~\ref{Fig:Loss convergence}, SecureBoost shows a similar learning curve with other non-federated methods on the training dataset and even performs slightly better than others on the test dataset. In addition, the convergence behavior of training and test loss of SecureBoost are very much alike GBDT and XGBoost. 

Next, to investigate how maximum depth of individual trees affects the runtime of learning, we vary the maximum depth of individual tree among $\left \{3,4,5,6,7,8\right \}$ and report the runtime of one boosting stage. As depicted in Figure~\ref{Fig:Scalability Analysis} (a), the runtime increases almost linearly with the maximum depth of individual trees, which indicates that we can train deep trees with relatively little additional time, which is very appealing in practice, especially in scenarios like big data.

Finally we study the impact of data size on the runtime of our proposed system. We augment the feature sets by feature products. We fix the maximum depth of individual regression trees to $3$ and vary the feature number in $\left \{50,500,1000,5000\right \}$ and the sample number in $\left \{5000,10000,30000\right \}$. We compare the runtime of one boosting stage to investigate how each variant affects the efficiency of the algorithm. We make similar observations on both Figure~\ref{Fig:Scalability Analysis} (b) and Figure~\ref{Fig:Scalability Analysis} (c), which imply that sample and feature numbers contribute equally to running time. In addition, we can see that our proposed framework scales well even with relatively big data.

\subsection{Performance of RL-SecureBoost}
To investigate the performance of RL-SecureBoost in both security and prediction accuracy, we aim to answer the following questions: (1) Does the first tree, built upon only features held by active party, learns enough information to reduce information leakage? (2) Does RL-SecureBoost suffer from a loss of acccuracy compared with SecureBoost?

First, we study the performance of RL-SecureBoost in security. Following the analysis in Section~\ref{sec:Security Discussion}, we evaluate information leakage in terms of leaf purity. Also, we know that as the leaf purity in the first tree increases, leaked information is reduced. Thereby, to verify the security of RL-SecureBoost, we have to illustrate that the first tree of RL-SecureBoost perform well enough to reduce the information leaked from the second tree. As shown in Table~\ref{table:Security}, we compare the mean leaf purity of the first tree with the second tree. In particular, the mean leaf purity is the weighted average, which is calculated by $\sum_{i=0}^{k}\frac{n_{i}}{n} p_{i}$. Here, $k$ and $n$ represents number of leaves and number of instances in total. $p_{i}$ and $n_{i}$ are defined as leaf purity and number of instances associated with leaf $i$. According to Table~\ref{table:Security}, the mean leaf purity decreases significantly from the first to the second tree on both datasets, which reflects a great reduction in information leakage. Moreover, the mean leaf purity of the second tree is just over $0.6$ on both datasets, which is good enough to ensure a safe protocol.

Next, to investigate the prediction performance of RL-SecureBoost, we compare RL-SecureBoost with SecureBoost with respect to the the first tree's performance and the overall performance. We consider commonly used metrics including accuracy, Area under ROC curve (AUC) and f1-score. The results are presented in Table~\ref{table:Accuracy}. As observed, RL-SecureBoost performs equally well compared to SecureBoost in almost all cases. We also conduct a pairwise Wilcoxon signed-rank test between RL-SecureBoost and SecureBoost. The comparison results indicate that RL-SecureBoost is as accurate as SecureBoost, with a significance level of $0.05$. The property of lossless is still guaranteed for RL-SecureBoost.

\section{Conclusion}\label{sec:Conclusion}
In this paper, we proposed a lossless privacy-preserving tree boosting algorithm, SecureBoost, to train a high-quality tree boosting model with private data split across multiple parties. We theoretically prove that our proposed framework is as accurate as non-federated gradient tree boosting counterparts. In addition, we analyze information leakage during the protocol execution and propose provable ways to reduce it. 

\section*{Acknowledgment}
This work was partially supported by the National Key Research and Development Program of China under Grant No. 2018AAA0101100. 

\ifCLASSOPTIONcaptionsoff
  \newpage
\fi



%
\bibliographystyle{IEEEtran}
\bibliography{reference}

\end{document}